\documentclass[letterpaper]{article} % DO NOT CHANGE THIS
\usepackage[preprint]{aaai2027}  % AAAI-style arXiv preprint
\usepackage[hyphens]{url}  % DO NOT CHANGE THIS
\usepackage{graphicx} % DO NOT CHANGE THIS
\usepackage{natbib}  % DO NOT CHANGE THIS AND DO NOT ADD ANY OPTIONS TO IT
\usepackage{caption} % DO NOT CHANGE THIS AND DO NOT ADD ANY OPTIONS TO IT
\usepackage{algorithm}
\usepackage{algorithmic}
\usepackage{newfloat}
\usepackage{listings}
\DeclareCaptionStyle{ruled}{labelfont=normalfont,labelsep=colon,strut=off} % DO NOT CHANGE THIS
\floatstyle{ruled}
\newfloat{listing}{tb}{lst}{}
\floatname{listing}{Listing}
\usepackage{booktabs}
\usepackage{multirow}

\title{EvoWiki: Incremental State Overwriting and Traceable Question Answering for Cross-Meeting Knowledge Evolution}
\author{Dongsheng Chen, Tianyu Wang, Wenhui Que\textsuperscript{*}}
\affiliations{
WeChat, Tencent Inc., Beijing, China\\
\texttt{\{joeydschen, tianyuwang, victorque\}@tencent.com}
}

\begin{document}
\maketitle

\begin{abstract}
In long-term collaboration spanning multiple meetings, factual states such as decisions, risks, and ownership are continually revised, overturned, and replaced. Existing long-context methods typically stack the entire history, while many RAG, LLM-Wiki, and structured-memory methods organize knowledge as static or append-only facts and rely on semantic relevance at read time. Without explicit modeling of intra-meeting decision processes and knowledge lifecycles, these approaches may retain conflicting old and new states simultaneously or discard history when updating snapshots, leading to stale retrieval and answers that are difficult to verify. We present \textbf{EvoWiki (Evolving Wiki)}, an incremental question-answering architecture for dynamic long-form text. EvoWiki decouples offline incremental construction (\textsc{Build}) from online structured reading (\textsc{Read}). \textsc{Build} captures the intra-meeting micro-evolution from proposal through discussion to decision and uses entity version chains and a fine-grained State-Overwrite Protocol to explicitly distinguish current valid states from superseded history while preserving meeting-level provenance anchors. \textsc{Read} bypasses relevance-based Top-$k$ retrieval over raw meetings and performs deterministic entity addressing, temporal resolution, and cross-entity multi-hop aggregation over the complete Wiki to produce grounded and traceable answers. We further introduce \textbf{CrossMeet}, a high-fidelity bilingual benchmark derived from real-world business seeds and designed to simulate long-term state evolution, covering factual consistency, temporal reasoning, and cross-meeting multi-hop reasoning. Across six datasets and two reader models, EvoWiki improves macro-average Judge Accuracy over the strongest baselines by 9.72 and 10.00 percentage points, respectively. Further analyses and human evaluation show that EvoWiki is more robust and factually faithful under frequent state flips, validating valid-state-oriented evolutionary reading as a more reliable technical approach to cross-meeting knowledge evolution.
\end{abstract}

% \noindent\textbf{Keywords:} cross-meeting question answering; incremental knowledge bases; state overwriting; temporal reasoning

\section{Introduction}

Successive meeting records characterize evolving project states: key decisions, risks, and responsibilities may be repeatedly revised, and historically valid statements may no longer represent current decisions. Cross-meeting QA must therefore identify the version valid at the query time and trace it to its source meeting. Prior work exposes the temporal-awareness limitations of static models and the challenges of meeting understanding, long-context QA, and cross-session knowledge updates \citep{li2026static,prasad2023meetingqa,thonet2025elitr,wu2024longmemeval}, but existing tasks still inadequately cover role binding, version replacement, and multi-hop evidence composition across meetings.

Three approaches remain insufficient. Long-context models place meeting histories in a single window, but nominal length does not guarantee reliable evidence use or reasoning \citep{hsieh2024ruler,yen2025helmet,modarressi2025nolima}. Conventional RAG \citep{lewis2020retrieval} ranks evidence by semantic relevance rather than validity; even recency- and conflict-aware methods \citep{vu2024freshllms,wang2025astute} may assign similar relevance to obsolete and current statements about the same entity. LLM-Wiki and structured-memory methods improve knowledge organization \citep{sarthi2024raptor,edge2024local,gutierrez2025rag,chen2026you} but do not model the lifecycles and replacement relations of cross-meeting decisions. Superseded and current states may therefore coexist and cause stale retrieval.

We therefore propose EvoWiki (Evolving Wiki). \textsc{Build} captures proposal--discussion--decision evolution and maintains versions and provenance through write-time coreference resolution, entity routing, and state overwriting; \textsc{Read} deterministically reads valid states from the complete Wiki without accessing raw meetings. This asymmetry moves disambiguation and conflict resolution to write time. We also introduce bilingual CrossMeet. Across six datasets and two readers, EvoWiki surpasses the strongest baselines by 9.72 and 10.00 percentage points, with additional analyses confirming robustness and factual faithfulness. The dataset and implementation will be released upon acceptance. Our contributions are: (1) We propose \textbf{EvoWiki}, whose asymmetric \textsc{Build}--\textsc{Read} design combines intra-meeting evolution, entity version chains, and fine-grained state overwriting to maintain a current valid view with traceable history and evidence. (2) We construct \textbf{CrossMeet}, a bilingual cross-meeting benchmark covering factual consistency, temporal reasoning, and multi-hop QA; each language contains 100 projects, 500 high-fidelity simulated meetings, and 2,000 QA pairs with average contexts exceeding 26K tokens, plus question-type, reasoning-hop, and cross-meeting evidence-chain annotations. (3) We enable traceable valid-state reasoning with low hallucination risk: experiments across six benchmarks and two readers, together with state-flip analysis and human evaluation, show reliable valid-state reading and factually faithful answers with verifiable evidence.

\begin{figure*}[t]
    \centering
    \includegraphics[width=\textwidth]{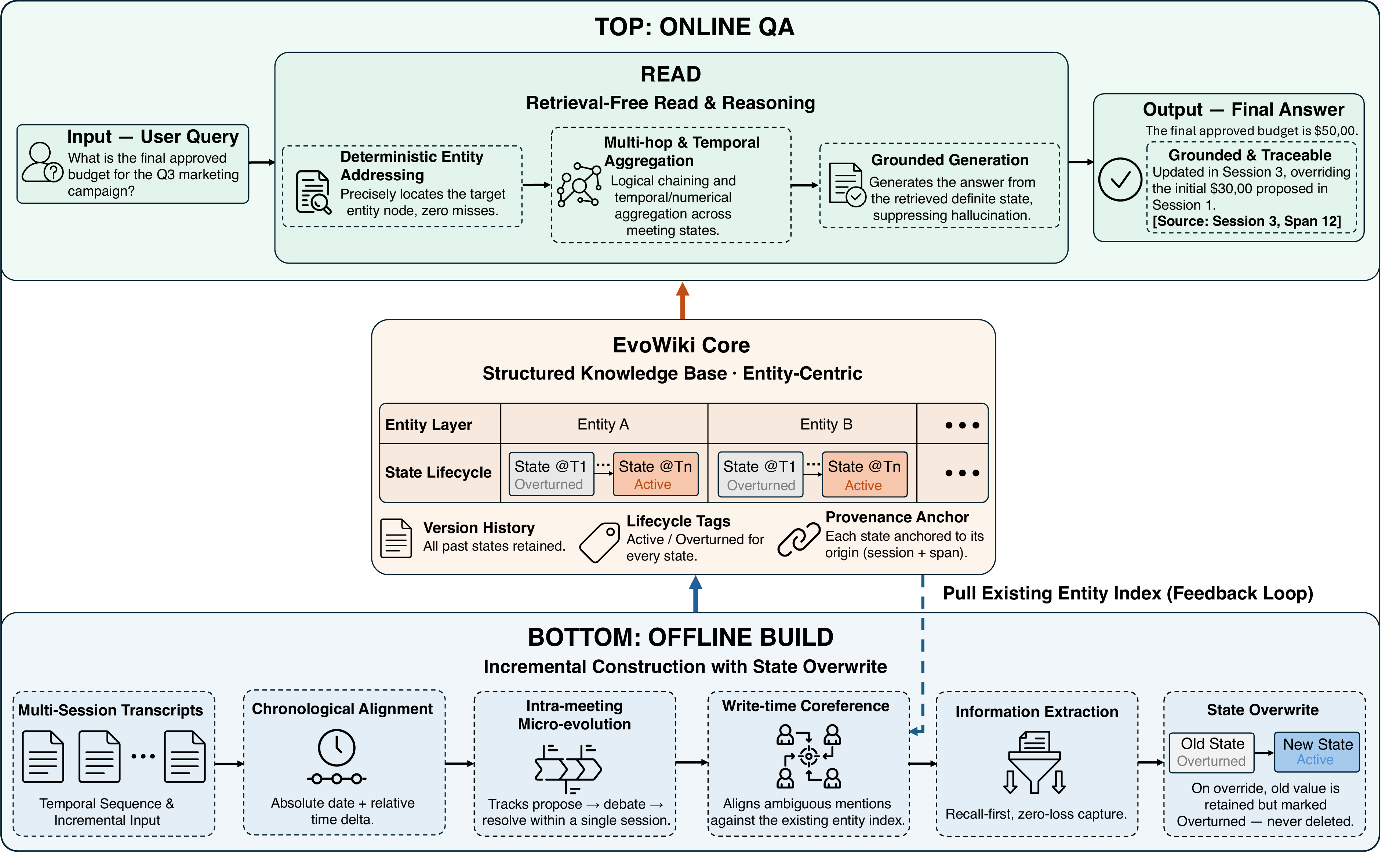}
    \caption{EvoWiki architecture. \textsc{Build} writes meetings chronologically into an entity-centered core with version and provenance information. Wiki-only \textsc{Read} uses deterministic addressing and cross-entity temporal aggregation to produce a grounded answer while retaining state-level provenance for evidence-trail recovery.}
    \label{fig:framework}
\end{figure*}

\section{Related Work}

\subsection{Long-Context QA, Long-Term Memory, and Meetings}

Long-context models and RAG offer complementary performance--cost profiles \citep{li2024retrieval}, yet nominal window length does not guarantee reliable evidence use \citep{bai2025longbench,hsieh2024ruler,yen2025helmet,modarressi2025nolima}. Prior work covers conversational evaluation and temporal or structured agent memory \citep{wu2024longmemeval,su2026beyond,latimer2026hindsight,jiang2026magma}, meeting understanding, QA, and generation \citep{prasad2023meetingqa,thonet2025elitr,zhu2025mfinmeeting,kirstein2025you}, and general multi-hop reasoning \citep{trivedi2022musique}. Most tasks focus on single meetings, general memory, or static evidence; CrossMeet targets state evolution, role binding, and multi-hop evidence across bilingual meeting sequences.

\subsection{Retrieval-Augmented Generation and Structured Knowledge}

Classical RAG accesses external memory through sparse, dense, or hybrid retrieval \citep{lewis2020retrieval,robertson2009probabilistic,karpukhin2020dense,zhang2025qwen3}; Self-RAG, Corrective RAG, and Adaptive-RAG improve retrieval control \citep{asai2024self,yan2024correctiveretrievalaugmentedgeneration,jeong2024adaptive}. WiCER compiles sources into persistent LLM-Wiki memory \citep{huerta2026wicer}. RAPTOR, GraphRAG/LightRAG, and HippoRAG~2/LogicRAG respectively use hierarchical summaries, entity graphs, and associative or multi-step reasoning \citep{sarthi2024raptor,edge2024local,guo2024lightrag,gutierrez2025rag,chen2026you}; other work studies adaptive structures, graph expansion, and path pruning \citep{li2025structrag,zhu2025knowledge,chen2026pathrag}. These methods do not explicitly maintain cross-meeting decision lifecycles and replacement relations. EvoWiki maintains both a current valid view and traceable version history at write time.

\subsection{Temporal Updating, Traceability, and Evaluation}

Temporal retrieval handles time-sensitive and arriving knowledge \citep{vu2024freshllms,zhang2024mrag,schumacher2025raster,hou2025synapticrag,liska2022streamingqa}. Parametric editing studies fact localization, multi-hop consistency, and lifelong updates \citep{meng2022locating,zhong2023mquake,wang2024wise,fang2025alphaedit,cheng2025serial}, but struggles to preserve meeting-level provenance; existing evaluation also examines factual retrieval and reasoning \citep{krishna2025fact}. EvoWiki instead binds current and superseded states to evidence in a versioned ledger, jointly supporting valid-state QA and historical audit. Unlike temporal knowledge graphs or append-only event sourcing, it processes unstructured meeting streams and maintains entity lifecycles at write time.

\section{Methodology}

\subsection{Task Formulation}

Given a chronologically ordered meeting sequence
\begin{equation}
\mathcal{D}=\{M_1,M_2,\ldots,M_T\},
\end{equation}
the goal is to answer query $q$ after processing the $T$-th meeting. Unlike static multi-document QA, an entity's attributes may be supplemented, replaced, or overturned by later meetings; the system must identify the state valid at the query time while retaining the historical evidence for its evolution.

EvoWiki represents atomic meeting knowledge as
\begin{equation}
s_i=\langle e_i,r_i,v_i,\tau_i,\ell_i,p_i\rangle,
\end{equation}
where $e_i,r_i,v_i,\tau_i,\ell_i$ denote the canonical entity, attribute or relation, state value, effective time, and lifecycle label, respectively, while $p_i=(m_i,\mathrm{span}_i)$ denotes the source meeting and evidence span. The system maintains a structured Wiki $\mathcal{W}_t$ under the causal constraint
\begin{equation}
\mathcal{W}_t=F_{\mathrm{Build}}(\mathcal{W}_{t-1},M_t),
\end{equation}
so processing $M_t$ accesses only the prior Wiki and the current meeting, never future meeting information.

\subsection{\textsc{Build}--\textsc{Read} Architecture}

Figure~\ref{fig:framework} presents EvoWiki's three components. Offline \textsc{Build} performs temporal alignment, intra-meeting micro-evolution parsing, write-time coreference resolution, information extraction, entity routing, and state overwriting. EvoWiki Core stores entities, active states, complete version histories, lifecycle labels, and provenance anchors. Online \textsc{Read} uses the complete Wiki as its sole context and generates an answer through deterministic entity addressing, cross-entity multi-hop reasoning, and temporal aggregation; provenance anchors bound to supporting states enable evidence-trail recovery. This read/write asymmetry moves temporal alignment, entity disambiguation, and conflict resolution to write time, lets queries reuse the construction result, and transforms online QA from re-identifying valid facts in conflicting passages into reading valid versions from a normalized state space.

\subsection{Offline Incremental Construction}

For each meeting $M_t$, \textsc{Build} executes four steps. First, temporal alignment (\textsc{Resolve}) normalizes the meeting timeline from dates and relative time expressions and parses the intra-meeting micro-evolution from proposal through discussion to decision, preventing candidate plans from being written as final states. Second, write-time coreference resolution (\textsc{Coref}) combines the current meeting with the entity index in $\mathcal{W}_{t-1}$ to map expressions such as ``the client,'' ``the above risk,'' or role titles to canonical entities. Fact extraction (\textsc{Extract}) then produces candidate updates, and entity routing (\textsc{Route}) binds them to the corresponding entity--attribute slots. Finally, the State-Overwrite Protocol merges the update set $\mathcal{U}_t$ into the Wiki:
\begin{equation}
\mathcal{U}_t=\mathrm{Route}(\mathrm{Extract}(\mathrm{Coref}(\mathrm{Resolve}(M_t),\mathcal{W}_{t-1}))),
\end{equation}
\begin{equation}
\mathcal{W}_t=\mathrm{Overwrite}(\mathcal{W}_{t-1},\mathcal{U}_t).
\end{equation}
Write-time coreference resolution and entity routing consolidate cross-meeting aliases, job titles, and elliptical expressions into one entity version chain, enabling strong role binding.

\subsection{State Overwriting and the Structured Wiki}

EvoWiki represents its knowledge base as
\begin{equation}
\mathcal{W}_t=(\mathcal{E}_t,\mathcal{H}_t,\mathcal{A}_t,\mathcal{P}_t),
\end{equation}
whose components denote entities and relations, complete version histories, the current active-state view, and provenance anchors. For each entity--attribute pair $(e,r)$, the system maintains a chronologically ordered version chain
\begin{equation}
\mathcal{H}_{e,r}^{(t)}=\langle s_{e,r}^{1},s_{e,r}^{2},\ldots,s_{e,r}^{n}\rangle.
\end{equation}
When a new state replaces, revises, or revokes the current active state, the old record is not physically deleted. The protocol closes its validity interval, labels it \texttt{Overturned}, appends the new state with an \texttt{Active} label, and creates an explicit version-replacement edge. Facts that do not conflict with the attribute remain in their respective slots. For any fixed entity--attribute pair, the protocol maintains at most one active version, where $\mathbf{1}[\cdot]$ denotes the indicator function:
\begin{equation}
\sum_{s\in\mathcal{H}_{e,r}^{(t)}}\mathbf{1}[\ell(s)=\mathrm{Active}]\leq 1.
\end{equation}
The current state therefore has a unique canonical entry point, while superseded versions and their provenance remain available for historical queries and audit. If a state's validity interval is $I(s)=[\tau_{\mathrm{start}},\tau_{\mathrm{end}})$, a query targeting time $\tau_q$ reads the version satisfying $\tau_q\in I(s)$; a current-state query without an explicit time defaults to the active version.

\subsection{Wiki-Only Online Reading}

During \textsc{Read}, the complete Wiki is serialized as the sole knowledge context:
\begin{equation}
C_T=\mathrm{Serialize}(\mathcal{W}_T),\qquad
(\hat{y},\hat{\mathcal{S}}_q)=G(q\mid C_T),
\end{equation}
where $\hat{y}$ is the answer and $\hat{\mathcal{S}}_q$ contains its supporting Wiki-state identifiers. Only $\hat{y}$ is evaluated; $\hat{\mathcal{S}}_q$ remains metadata for evidence-trail recovery via the traceability mechanism described below. \textsc{Read} is isolated from raw meetings: it performs no source-text Top-$k$ retrieval and has no fallback that accesses raw meetings when the context window permits. Online inference locates valid states, performs multi-hop or temporal aggregation along entity relations and version timelines, and generates answers from those states and their provenance anchors. ``Deterministic Entity Addressing'' in Figure~\ref{fig:framework} denotes reading valid states in the Wiki by entity, attribute, and lifecycle label rather than retrieving raw meetings or an external corpus. Because \textsc{Build} has written valid states into a unified structure, \textsc{Read} bypasses source-text relevance ranking and Top-$k$ errors without deciding the current version among mixed old and new passages.

\subsection{Traceability and Cost}

Every state is linked to its source meeting and evidence span, while version-replacement edges record how states evolve. For the supporting-state set $\hat{\mathcal{S}}_q$ returned with an answer, its evidence chain is
\begin{equation}
\mathrm{Trace}(\hat{y})=\bigcup_{s\in\hat{\mathcal{S}}_q}p(s)\ \cup\ \mathrm{VersionEdges}(\hat{\mathcal{S}}_q).
\end{equation}
The system can therefore locate the meeting evidence supporting the final answer and trace why one state replaced an earlier version. If all meetings produce $U$ state updates, retaining all historical versions requires $O(U)$ storage. For a serialized Wiki of length $S_W$, each query has an input-context size of $O(S_W)$ and makes no retrieval call to an external corpus. The main \textsc{Build} cost occurs during offline writing and can be amortized over subsequent queries; correspondingly, complete-Wiki input length grows linearly with $S_W$. This linear reading cost is an explicit architectural trade-off for broader state coverage and reduced risks of retrieval omission and obsolete-version selection.

\section{Experimental Setup}

\subsection{CrossMeet Construction and Quality Validation}

CrossMeet follows a controlled project-blueprint--meeting-sequence--cross-meeting-QA pipeline. Its English and Chinese versions are generated independently rather than translated from one another. To improve the business realism of the simulated meetings, project blueprints are expanded from abstract project topics representing common workflows on large-scale digital platforms. Only generic topic categories and collaboration patterns are used, without any real business records, personal identifiers, operational metrics, or sensitive business content. Claude Opus 4.5 \citep{anthropic2025opus45} constructs the meetings and QA instances; Gemini 3.1 Pro \citep{deepmind2026gemini31}, Claude Opus 4.7 \citep{anthropic2026opus47}, and DeepSeek-V4-Pro \citep{xu2026deepseek} cross-validate answer--evidence consistency, answerability, and question-type labels and independently classify question types to compute Fleiss' $\kappa$ as agreement over the three categories.

\begin{table}[t]
    \centering
    \small
    \resizebox{\columnwidth}{!}{
        \begin{tabular}{@{} l r r @{}}
            \toprule
            \textbf{Metric} & \textbf{CM-EN} & \textbf{CM-ZH} \\
            \midrule
            Topics                           & 100    & 100    \\
            Meetings                         & 500    & 500    \\
            QA Pairs                         & 2,000  & 2,000  \\
            Context Tokens (Avg.)            & 26,861 & 29,480 \\
            Answer Tokens (Avg.)             & 64.6   & 108.1  \\
            Reasoning Hops (Avg. / Max.) & 2.84 / 5 & 2.77 / 5 \\
            Taxonomy $\kappa$                & 0.937  & 0.909  \\
            \bottomrule
        \end{tabular}
    }
    \caption{Statistics of CrossMeet-EN and CrossMeet-ZH. Taxonomy $\kappa$ measures question-type classification agreement among three validation models.}
    \label{tab:dataset_statistics_transposed}
\end{table}

\begin{table}[t]
    \centering
    \small
    \resizebox{\columnwidth}{!}{
        \begin{tabular}{@{} ll cc @{}}
            \toprule
            \textbf{Dimension} & \textbf{Type} & \textbf{CM-EN} & \textbf{CM-ZH} \\
            \midrule
            \multirow{3}{*}{Question} 
            & Factual Consistency & 1000 (50.0\%) & 1000 (50.0\%) \\
            & Timeline Reasoning  & 600 (30.0\%)  & 600 (30.0\%) \\
            & Multi-hop Reasoning & 400 (20.0\%)  & 400 (20.0\%) \\
            \midrule
            \multirow{4}{*}{Hops} 
            & 2 Hops & 941 (47.0\%) & 1005 (50.2\%) \\
            & 3 Hops & 635 (31.8\%) & 608 (30.4\%) \\
            & 4 Hops & 235 (11.8\%) & 222 (11.1\%) \\
            & 5 Hops & 189 (9.4\%)  & 165 (8.2\%) \\
            \bottomrule
        \end{tabular}
    }
    \caption{Distribution of question types and reasoning hops in CrossMeet.}
    \label{tab:question_hop_distribution}
\end{table}

Table~\ref{tab:dataset_statistics_transposed} summarizes dataset scale and quality. Each language contains 100 projects, 500 consecutive meetings, and 2,000 QA pairs, with average context lengths of 26,861 and 29,480 tokens for English and Chinese, respectively. Fleiss' $\kappa$ for the three-model question-type classification is 0.937 in English and 0.909 in Chinese, indicating high agreement in the task taxonomy. Table~\ref{tab:question_hop_distribution} reports question-type and hop distributions: factual-consistency, temporal-reasoning, and cross-meeting multi-hop questions account for 50\%, 30\%, and 20\%. Every question requires at least two hops; four- and five-hop instances constitute 21.2\% of the English data and 19.3\% of the Chinese data, ensuring long-evidence-chain coverage.

\paragraph{Independent human validation.}
We sample 150 QA instances and their complete state-evolution paths from each CrossMeet language, yielding 300 of 4,000 instances. Three independent annotators with natural language processing experience, none involved in data generation, review every sample without access to generation prompts or automatic validation labels. Each receives the question, reference answer, cited evidence, and complete evolution path. They assess whether the answer is supported, whether proposal--discussion--decision transitions are authentic and coherent, and whether question-type and reasoning-hop annotations are correct. Each annotator judges the samples independently; pass rates follow majority decisions, and inter-annotator agreement is measured with Fleiss' $\kappa$. Table~\ref{tab:appendix_dataset_audit} summarizes the results.

\begin{table}[t]
    \centering
    \small
    \resizebox{\columnwidth}{!}{
        \begin{tabular}{@{} l c c @{}}
            \toprule
            \textbf{Evaluation Dimension} & \textbf{Pass Rate} & \textbf{Fleiss' $\kappa$} \\
            \midrule
            Answer--Evidence             & 98.3\% & 0.89 \\
            Meeting/Update Coherence     & 95.7\% & 0.84 \\
            Type/Hop Accuracy            & 97.7\% & 0.91 \\
            \bottomrule
        \end{tabular}
    }
    \caption{Human validation on 300 CrossMeet instances sampled equally from English and Chinese.}
    \label{tab:appendix_dataset_audit}
\end{table}

All dimensions exceed a 95\% pass rate, with agreement of at least 0.84. The 98.3\% answer--evidence rate confirms that reference answers are grounded in the supplied meeting evidence. The meeting/update result further indicates that the simulated discussions preserve credible proposal, revision, rejection, and resolution processes rather than merely presenting disconnected facts. Metadata accuracy shows that the benchmark's question categories and annotated reasoning depth are also reliable. This jointly validates both reasoning targets and metadata used in stratified analyses. The audit thus complements automatic cross-validation with direct human evidence of benchmark fidelity and annotation quality.

\subsection{Evaluation Datasets}

In addition to CrossMeet-EN and CrossMeet-ZH, experiments use four public benchmarks. MeetingQA is based on human-recorded AMI scenario meetings and evaluates QA over meeting transcripts \citep{prasad2023meetingqa}; ELITR-Bench is based on ASR transcripts of real project meetings and covers retrieval, summarization, and QA in long meetings \citep{thonet2025elitr}; LongMemEval tests long-term memory and knowledge updates across sessions \citep{wu2024longmemeval}; and MuSiQue evaluates static compositional multi-hop reasoning \citep{trivedi2022musique}. MeetingQA and ELITR-Bench both use human meeting transcripts as their core data. Together, the six datasets cover bilingual cross-meeting evolution, single-meeting understanding, long-term interactive memory, and general multi-hop reasoning.

\subsection{Baselines, Reader Models, and Metrics}

We compare nine representative baselines: Direct LLM for full-context reading \citep{li2024retrieval}; VanillaRAG (BM25), VanillaRAG (Dense), and VanillaRAG (Hybrid) for sparse, dense, and fused retrieval \citep{robertson2009probabilistic,karpukhin2020dense,zhang2025qwen3}; RAPTOR for hierarchical summarization \citep{sarthi2024raptor}; GraphRAG \citep{edge2024local} and LightRAG \citep{guo2024lightrag} for graph retrieval; HippoRAG~2 for associative memory \citep{gutierrez2025rag}; and LogicRAG for query-time logical decomposition \citep{chen2026you}. Dense retrieval uses Qwen3-Embedding-8B \citep{zhang2025qwen3}. Together, they span direct reading, sparse/dense retrieval, hierarchical compression, graph retrieval, associative memory, and logical planning. All retrieval baselines share tiered chunking and budgets scaled to dataset context length; Direct LLM reads the full raw context within each reader's native window.

EvoWiki's offline \textsc{Build} stage uniformly uses DeepSeek-V4-Flash \citep{xu2026deepseek} to perform temporal alignment, intra-meeting micro-evolution parsing, write-time coreference resolution, fact extraction, and entity routing, and to produce the structured Wiki according to the State-Overwrite Protocol. Strong structured baselines such as GraphRAG and HippoRAG~2 likewise strictly follow their officially recommended complete offline LLM-based graph construction and indexing pipelines. Thus, all methods start from the same raw meetings and retain their native construction and reading pipelines: the main results compare end-to-end system capability, while the matched ablation study identifies the contributions of state overwriting and structural tags. Online \textsc{Read} uses either DeepSeek-V4-Flash \citep{xu2026deepseek} or Qwen3.5-397B-A17B \citep{team2026qwen3} as the reader; both settings share the same \textsc{Build} configuration to isolate reader-model differences. EvoWiki strictly follows Wiki-only \textsc{Read}: the online reader receives only the complete Wiki produced by \textsc{Build} and never accesses raw meeting records. Under the same reader model, all methods use identical questions, prompts, and generation settings. We do not provide conventional RAG with EvoWiki's intermediate structures because doing so would alter its native raw-text retrieval paradigm and make it dependent on EvoWiki's extractor, rather than compare each method's own end-to-end knowledge-organization capability.

Judge Accuracy is the primary metric. Following the LLM-as-a-Judge paradigm \citep{zheng2023judging}, Gemini 3.1 Pro \citep{deepmind2026gemini31} assigns 0 to an incorrect answer, 0.5 to a correct but incomplete core conclusion, and 1 to a fully correct answer. Each method runs five times with matched online decoding while its offline Wiki, graph, or index remains fixed; Table~\ref{tab:main_results} reports mean scores from this sole primary judge. Scores are averaged per dataset and then equally macro-averaged across all six. We also report abstention and anonymized pairwise human win/tie/loss results, and analyze ablations, state flips, evidence positions, error attribution, and qualitative cases; the pairwise results are independently judged by human evaluators.

\paragraph{Retrieval and decoding configurations.}
All retrieval baselines use a unified length-tiered configuration determined solely by each dataset's median context length rather than tuned on experimental results. MeetingQA and MuSiQue use a chunk size of 200 tokens, an overlap of 25 tokens, and Top-3 retrieval; CrossMeet-EN, CrossMeet-ZH, ELITR-Bench, and LongMemEval use a chunk size of 512 tokens, an overlap of 64 tokens, and Top-10 retrieval. RAPTOR halves the leaf-chunk configuration to 100/12 tokens for short-context datasets and 256/32 tokens for long-context datasets, while retaining Top-3 and Top-10 retrieval, respectively.

Except for the multi-run stability experiment, all methods use the same deterministic final-reader configuration: temperature is set to 0.0, the maximum generation length is 1,024 tokens, and explicit reasoning output is disabled. Top-p is not explicitly specified and therefore follows the default of the corresponding serving endpoint. The five-run stability analysis varies only the online-generation seed and temperature, using 0.1, 0.3, 0.5, 0.7, and 0.9; the offline Wiki, graph, or index remains fixed, and the maximum generation length and all other settings are unchanged.

Direct LLM performs no chunking, retrieval, or structural compression and instead feeds the complete raw context directly to the corresponding reader model. Qwen3.5-397B-A17B has a native context window of 262,144 tokens (256K), while DeepSeek-V4-Flash supports 1,048,576 tokens (1M); all raw experimental contexts fall within the input capacity of the corresponding model.

\section{Main Results}

\subsection{Overall Performance}

\begin{table*}[t]
    \centering
    \small
    \resizebox{\textwidth}{!}{
        \begin{tabular}{@{} ll *{7}{c} @{}}
            \toprule
            \textbf{Model} & \textbf{Method} & \textbf{CM-EN} & \textbf{CM-ZH} & \textbf{ELITR} & \textbf{LongMemEval} & \textbf{MeetingQA} & \textbf{MuSiQue} & \textbf{Avg} \\
            \midrule
            \multirow{10}{*}{\textbf{DeepSeek-V4-Flash}}
            & Direct LLM          & 65.90 & 66.03 & 59.62 & 19.80 & 37.82 & 49.03 & 49.70 \\
            & VanillaRAG (BM25)   & 43.98 & 54.93 & 48.46 & 37.60 & 37.40 & 45.92 & 44.72 \\
            & VanillaRAG (Dense)  & 51.10 & 59.43 & 55.38 & 34.60 & 37.12 & 46.13 & 47.29 \\
            & VanillaRAG (Hybrid) & 51.60 & 58.75 & 51.15 & 39.20 & 37.06 & 46.03 & 47.30 \\
            & RAPTOR              & 49.20 & 55.20 & 46.54 & 40.10 & 33.84 & 48.57 & 45.57 \\
            & GraphRAG            & 56.33 & 61.30 & 58.08 & 39.10 & 35.40 & 51.76 & 50.33 \\
            & HippoRAG2           & 51.60 & 59.20 & 56.15 & 39.70 & 36.17 & 52.32 & 49.19 \\
            & LightRAG            & 52.40 & 60.30 & 52.69 & 44.30 & 35.72 & 49.50 & 49.15 \\
            & LogicRAG            & 54.20 & 58.50 & 54.62 & 48.50 & 29.42 & 56.99 & 50.37 \\
            & EvoWiki (Ours)      & \textbf{68.78} & \textbf{69.55} & \textbf{60.77} & \textbf{61.60} & \textbf{39.16} & \textbf{60.67} & \textbf{60.09} \\
            \midrule
            \multirow{10}{*}{\textbf{Qwen3.5-397B-A17B}}
            & Direct LLM          & 71.43 & 75.13 & 63.46 & 22.20 & 39.92 & 45.28 & 52.90 \\
            & VanillaRAG (BM25)   & 44.98 & 61.18 & 54.62 & 42.20 & 40.13 & 42.72 & 47.64 \\
            & VanillaRAG (Dense)  & 52.45 & 65.63 & 59.62 & 41.00 & 39.76 & 44.06 & 50.42 \\
            & VanillaRAG (Hybrid) & 53.35 & 65.45 & 58.85 & 43.20 & 40.33 & 43.42 & 50.77 \\
            & RAPTOR              & 50.20 & 60.30 & 53.46 & 43.30 & 36.33 & 48.14 & 48.62 \\
            & GraphRAG            & 57.05 & 65.23 & 58.08 & 47.40 & 36.11 & 52.15 & 52.67 \\
            & HippoRAG2           & 52.65 & 64.50 & 59.23 & 41.90 & 39.68 & 51.65 & 51.60 \\
            & LightRAG            & 53.20 & 66.20 & 58.46 & 46.30 & 37.02 & 53.43 & 52.44 \\
            & LogicRAG            & 55.35 & 67.83 & 59.62 & 51.90 & 28.53 & 54.90 & 53.02 \\
            & EvoWiki (Ours)      & \textbf{73.18} & \textbf{77.28} & \textbf{64.62} & \textbf{66.00} & \textbf{40.81} & \textbf{56.25} & \textbf{63.02} \\
            \bottomrule
        \end{tabular}
    }
    \caption{Judge Accuracy (\%) across six datasets. Best results within each reader block are shown in bold.}
    \label{tab:main_results}
\end{table*}

\begin{table*}[t]
    \centering
    \small
    \resizebox{\textwidth}{!}{
        \begin{tabular}{@{} l *{7}{c} @{}}
            \toprule
            \textbf{Variant} & \textbf{CM-EN} & \textbf{CM-ZH} & \textbf{ELITR} & \textbf{LongMemEval} & \textbf{MeetingQA} & \textbf{MuSiQue} & \textbf{Avg} \\
            \midrule
            w/o Overwrite  & 62.63 & 65.13 & 41.54 & 47.20 & 36.37 & 55.88 & 51.46 \\
            w/o Coref      & 61.03 & 66.60 & 43.46 & 43.20 & 35.74 & 55.52 & 50.93 \\
            w/o Entity     & 63.88 & 68.23 & 47.31 & 49.00 & 36.65 & 57.30 & 53.73 \\
            w/o Tags       & 60.43 & 62.78 & 42.31 & 41.40 & 35.50 & 56.29 & 49.79 \\
            EvoWiki        & \textbf{68.78} & \textbf{69.55} & \textbf{60.77} & \textbf{61.60} & \textbf{39.16} & \textbf{60.67} & \textbf{60.09} \\
            \bottomrule
        \end{tabular}
    }
    \caption{Component ablation of EvoWiki with DeepSeek-V4-Flash, reporting Judge Accuracy (\%) across six datasets.}
    \label{tab:ablation_results}
\end{table*}

Table~\ref{tab:main_results} summarizes the main results. With DeepSeek-V4-Flash, EvoWiki achieves the highest Judge Accuracy on all six datasets and a macro-average of 60.09, 9.72 percentage points above LogicRAG (50.37). With Qwen3.5-397B-A17B, its macro-average is 63.02, 10.00 points above LogicRAG (53.02). Consistent gains across readers indicate that the improvement stems from knowledge-state organization rather than a particular reader. On LongMemEval, EvoWiki scores 61.60 and 66.00, leading the strongest baselines by 13.10 and 14.10 points and confirming that state overwriting mitigates long-running knowledge conflicts and evidence-localization difficulty; its lead on the other five datasets shows that the advantage extends beyond CrossMeet.

\subsection{Ablation Study}

Table~\ref{tab:ablation_results} quantifies the contribution of each component under DeepSeek-V4-Flash. The w/o Overwrite variant keeps upstream temporal alignment, intra-meeting micro-evolution parsing, coreference resolution, fact extraction, entity routing, the entity-centered Wiki, and the reader unchanged; at write time, it discards Active labels emitted by the generic \textsc{Build} prompt, suppresses Overturned labels, and disables lifecycle transitions and version-replacement edges, causing conflicting states to coexist in an append-only form without lifecycle distinctions. It therefore provides a structured control with the same extraction strength as full EvoWiki but without version invalidation. The w/o Tags variant retains the same Wiki content but removes all structural tags used to organize entities, relations, time, states, and provenance, thereby measuring the overall effect of explicit structural representation. Full EvoWiki achieves an average accuracy of 60.09. Removing state overwriting, write-time coreference resolution, entity-centered organization, or all structural tags reduces the average to 51.46, 50.93, 53.73, and 49.79, corresponding to drops of 8.63, 9.16, 6.36, and 10.30 percentage points. The components play complementary roles: removing state overwriting reintroduces historical conflicts into the reading context; removing write-time coreference resolution causes entity fragmentation and broken evidence chains; and removing entity-centered organization increases the difficulty of cross-document addressing and aggregation. Removing all structural tags causes the largest drop (10.30 points), showing that entity, relation, temporal, state, and provenance boundaries jointly guide \textsc{Read}; without them, the reader must reconstruct structure and version validity from flattened content and is more likely to confuse current and superseded states.

\begin{table*}[!tb]
    \centering
    \small
    \resizebox{\textwidth}{!}{
        \begin{tabular}{@{} l c c c c c c c @{}}
            \toprule
            \textbf{Method} & \textbf{CM-EN} & \textbf{CM-ZH} & \textbf{ELITR} & \textbf{LongMemEval} & \textbf{MeetingQA} & \textbf{MuSiQue} & \textbf{Macro-Avg} \\
            \midrule
            Direct LLM  & $65.90\pm0.74$ & $66.03\pm0.68$ & $59.62\pm0.91$ & $19.80\pm1.22$ & $37.82\pm0.95$ & $49.03\pm0.83$ & $49.70\pm0.89$ \\
            VanillaRAG & $51.60\pm1.10$ & $58.75\pm1.05$ & $51.15\pm0.96$ & $39.20\pm1.18$ & $37.06\pm1.02$ & $46.03\pm0.93$ & $47.30\pm1.04$ \\
            GraphRAG   & $56.33\pm0.82$ & $61.30\pm0.75$ & $58.08\pm0.79$ & $39.10\pm1.06$ & $35.40\pm0.88$ & $51.76\pm0.81$ & $50.33\pm0.85$ \\
            LogicRAG   & $54.20\pm0.76$ & $58.50\pm0.72$ & $54.62\pm0.83$ & $48.50\pm0.97$ & $29.42\pm1.10$ & $56.99\pm0.78$ & $50.37\pm0.86$ \\
            EvoWiki    & $\mathbf{68.78\pm0.52}$ & $\mathbf{69.55\pm0.49}$ & $\mathbf{60.77\pm0.58}$ & $\mathbf{61.60\pm0.71}$ & $\mathbf{39.16\pm0.69}$ & $\mathbf{60.67\pm0.55}$ & $\mathbf{60.09\pm0.59}$ \\
            \bottomrule
        \end{tabular}
    }
    \caption{Performance stability across five runs under the DeepSeek-V4-Flash reader. Results are Judge Accuracy (\%) reported as mean $\pm$ standard deviation; VanillaRAG denotes the Hybrid variant.}
    \label{tab:appendix_variance}
\end{table*}

\begin{figure}[t]
    \centering
    \includegraphics[width=\columnwidth]{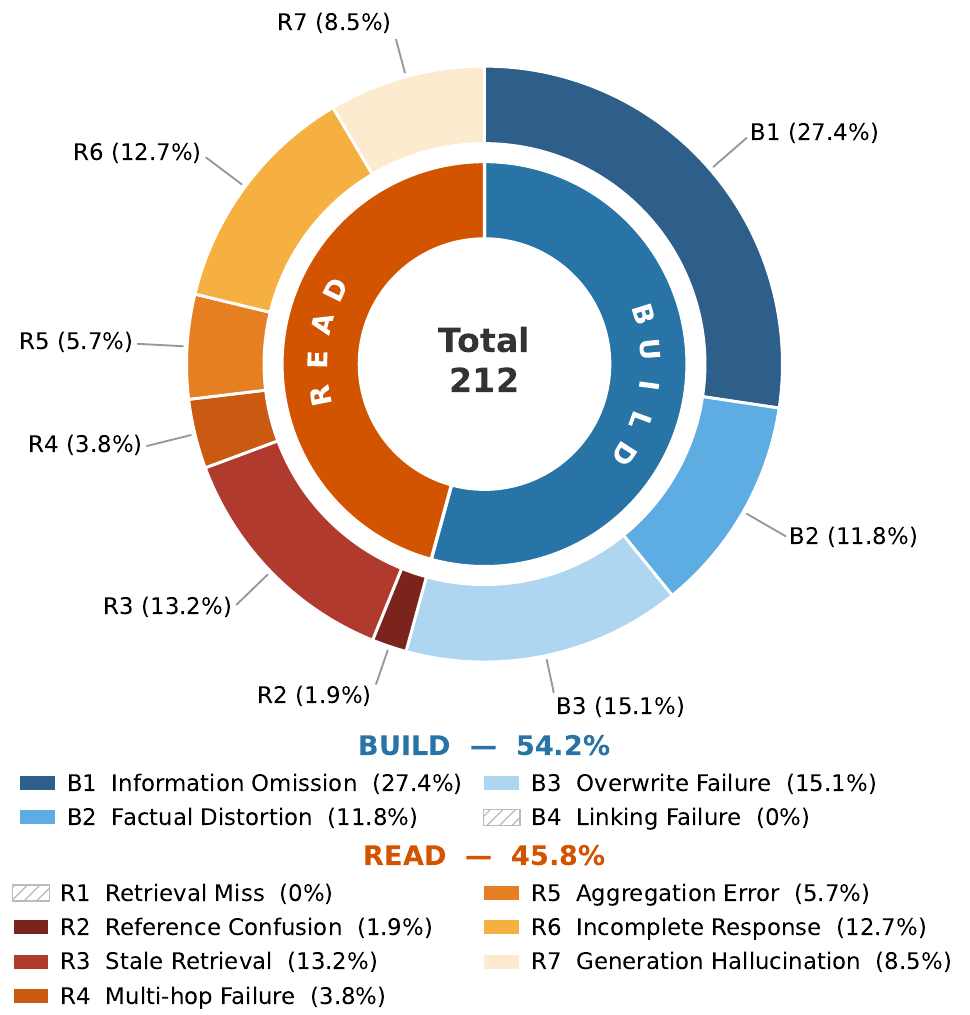}
    \caption{Error attribution for 212 EvoWiki errors, separated into \textsc{Build} and \textsc{Read} causes.}
    \label{fig:appendix_error_attribution}
\end{figure}

\begin{table}[t]
    \centering
    \small
    \resizebox{\columnwidth}{!}{
        \begin{tabular}{@{} l r r r @{}}
            \toprule
            \textbf{Method} & \textbf{Latency (s)} & \textbf{Read Tokens} & \textbf{Build Tokens} \\
            \midrule
            EvoWiki (Ours)          & 3.60 & 17,143 & 114,016 \\
            Direct LLM              & 6.54 & 26,374 & -- \\
            VanillaRAG (Hybrid)     & \textbf{1.01} & \textbf{3,481} & -- \\
            GraphRAG                & 1.94 & 9,452 & 131,298 \\
            LogicRAG                & 4.68 & 11,238 & -- \\
            LightRAG                & 1.78 & 7,352 & 67,263 \\
            \bottomrule
        \end{tabular}
    }
    \caption{Mean query latency and read/build token usage on CrossMeet-EN. Read costs are averaged per query and build costs per project.}
    \label{tab:appendix_read_efficiency}
\end{table}

\subsection{Performance Stability}

We compare EvoWiki with Direct LLM, VanillaRAG, GraphRAG, and LogicRAG under the DeepSeek-V4-Flash reader. Each method is run five times with different online-generation seeds and temperatures in $\{0.1,0.3,0.5,0.7,0.9\}$, while its offline Wiki, graph, or index is constructed once and held fixed across runs; Gemini 3.1 Pro scores all outputs. Table~\ref{tab:appendix_variance} reports mean Judge Accuracy and deviation across these decoding conditions. Because seed and temperature vary jointly, the deviation measures overall online-decoding robustness rather than seed-only variation. EvoWiki achieves the highest mean on all six datasets and the lowest macro-average deviation (0.59).

\section{Analysis and Discussion}

\subsection{Inference Efficiency and Construction Cost}

We measure QA efficiency on CrossMeet-EN with DeepSeek-V4-Flash as the reader, using 200 QA instances from 20 projects. Questions are processed serially, while project indexes, graphs, and Wiki structures are constructed, warmed, and cached before query timing. Table~\ref{tab:appendix_read_efficiency} reports non-streaming end-to-end \textsc{Read} latency, mean \textsc{Read} tokens per query, and mean one-time \textsc{Build} tokens per project; ``--'' denotes methods without offline construction. Latency is averaged across all instances and measures complete response time rather than time to first token. Compared with Direct LLM, EvoWiki reduces latency by 45.0\% and \textsc{Read}-token usage by 35.0\%. Retrieval methods remain cheaper online because they expose only a small Top-$k$ context, whereas EvoWiki reads the complete Wiki to preserve access to valid states and their histories. The measurements capture the efficiency--coverage trade-off, with the one-time \textsc{Build} cost amortized across subsequent queries.

\begin{table}[t]
    \centering
    \small
    \resizebox{\columnwidth}{!}{
        \begin{tabular}{@{} l c c c @{}}
            \toprule
            \textbf{Method} & \textbf{Gemini 3.1 Pro} & \textbf{Claude Opus 4.7} & \textbf{DeepSeek-V4-Pro} \\
            \midrule
            EvoWiki      & \textbf{60.50} & \textbf{60.17} & \textbf{59.83} \\
            Direct LLM   & 50.17 & 50.50 & 49.83 \\
            VanillaRAG   & 46.83 & 46.50 & 46.33 \\
            GraphRAG     & 49.83 & 50.00 & 49.50 \\
            LogicRAG     & 50.00 & 50.33 & 49.33 \\
            \bottomrule
        \end{tabular}
    }
    \caption{Judge Accuracy (\%) on the 300-question subset. Rows denote evaluated methods and columns denote judge models. Agreement with Gemini 3.1 Pro over 1,500 responses is $\kappa=0.901$ for Claude Opus 4.7 and $\kappa=0.887$ for DeepSeek-V4-Pro. VanillaRAG denotes the Hybrid variant.}
    \label{tab:appendix_judge_calibration}
\end{table}

\begin{figure}[t]
    \centering
    \includegraphics[width=\columnwidth]{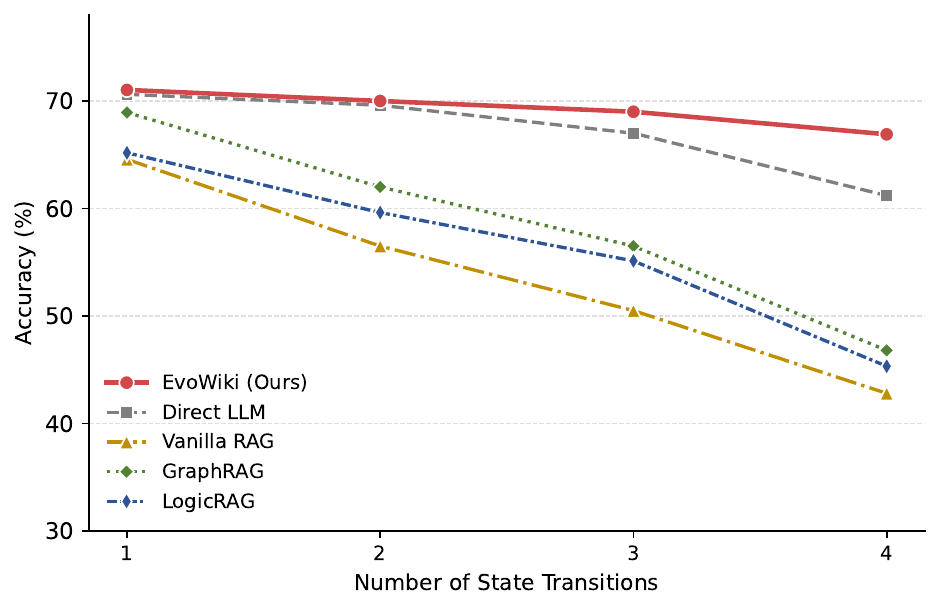}
    \caption{Judge Accuracy as the number of state flips increases on CrossMeet-EN.}
    \label{fig:state_flips}
\end{figure}

\begin{figure}[t]
    \centering
    \includegraphics[width=\columnwidth]{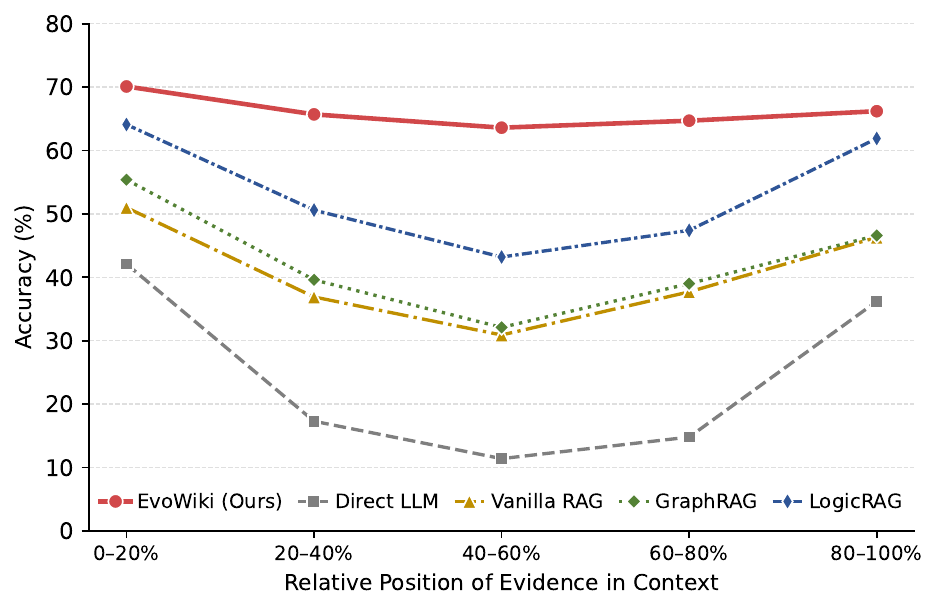}
    \caption{Judge Accuracy by the relative position of supporting evidence in LongMemEval.}
    \label{fig:position_effect}
\end{figure}

\begin{table*}[t]
    \centering
    \small
    \resizebox{\textwidth}{!}{
        \begin{tabular}{@{} p{0.15\linewidth} p{0.85\linewidth} @{}}
            \toprule
            \textbf{Source (Summary)} & The project's ``primary launch risk'' evolved twice across meetings. It was first flagged as an \textit{SOC2 certification delay}; in Meeting 4 it was updated to \textit{insufficient customer onboarding capacity} (only 2 customer support staff covering 3 enterprise accounts); in Meeting 5 this risk was overridden again to the \textbf{third customer's contract negotiation possibly slipping past Q4, affecting the ARR milestone}. The question asks for the primary launch risk at the delivery stage, i.e., it requires the latest (Meeting 5) state, not any overturned earlier risk. \\
            \midrule
            \textbf{Question}       & What was the primary launch risk identified for the Nexus Analytics project at the delivery stage? \\
            \midrule
            \textbf{Reference}      & Third customer contract negotiation may slip past Q4, affecting ARR milestone \\
            \midrule
            \textbf{EvoWiki}        & \textbf{Third customer contract negotiation potentially slipping past Q4, affecting ARR milestone.} \\
            \textbf{Direct LLM}     & \textit{Customer onboarding capacity.} (Landed on the Meeting-4 risk, one version behind---missed the Meeting-5 override.) \\
            \textbf{VanillaRAG}     & \textit{SOC2 certification delay.} (Landed on an even earlier risk that had already been overturned twice.) \\
            \textbf{GraphRAG}       & \textit{Customer onboarding capacity, with only 2 customer support staff for 3 enterprise accounts.} (Same stale Meeting-4 state as Direct LLM.) \\
            \textbf{LogicRAG}       & \textit{Unanswerable} (failed to locate any valid risk description and abstained.) \\
            \bottomrule
        \end{tabular}
    }
    \caption{Case study of cross-meeting launch-risk evolution and final-state recovery.}
    \label{tab:appendix_case_study_risk}
\end{table*}

\begin{figure*}[t]
    \centering
    \includegraphics[width=\textwidth]{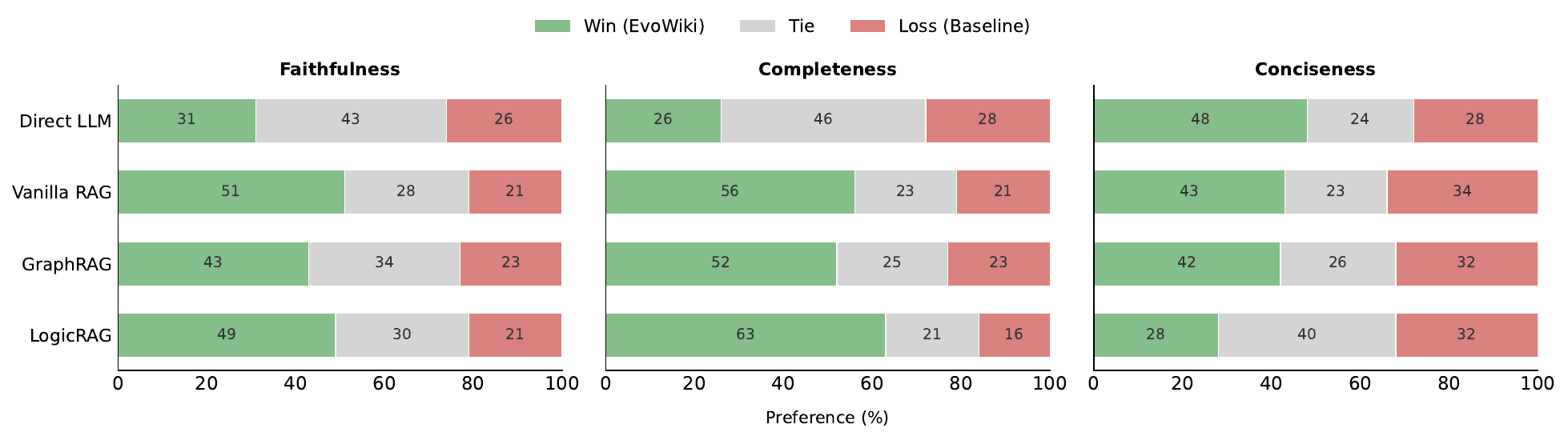}
    \caption{Pairwise human evaluation of EvoWiki against representative baselines on faithfulness, completeness, and conciseness. Results are percentages of wins, ties, and losses from EvoWiki's perspective.}
    \label{fig:human_eval_wtl}
\end{figure*}

\subsection{Cross-Model Judge Robustness}

We sample 50 questions from each benchmark to form a 300-question subset that preserves the equal-dataset weighting of the primary metric. Under the DeepSeek-V4-Flash reader, outputs from EvoWiki and four representative baselines yield 1,500 anonymized responses per judge. All three judges grade exactly the same predictions. Gemini 3.1 Pro, Claude Opus 4.7, and DeepSeek-V4-Pro independently apply the same 0/0.5/1 rubric using only the question, reference answer, and system response; method identities and other judges' decisions are hidden. Quadratic-weighted Cohen's $\kappa$ measures each auxiliary judge's agreement with Gemini over all responses and respects the ordinal distance between incorrect, partially correct, and fully correct ratings. Table~\ref{tab:appendix_judge_calibration} reports both system scores and cross-judge agreement. All three judges rank EvoWiki first. Its margins over the strongest baseline are 10.33, 9.67, and 10.00 points under Gemini, Claude, and DeepSeek, respectively. Claude and DeepSeek achieve agreement scores of 0.901 and 0.887 with Gemini, indicating that both the absolute assessments and method ordering remain stable across model families rather than reflecting one judge's preference. Because the comparison includes full-context reading, conventional retrieval, static graph organization, and logical graph reasoning, the conclusion is robust to both judge origin and baseline paradigm.

\subsection{Error Attribution}

Figure~\ref{fig:appendix_error_attribution} divides 212 EvoWiki errors into \textsc{Build} (115, 54.2\%) and \textsc{Read} (97, 45.8\%) failures. The main failure modes are extraction omission (27.4\%), overwrite failure (15.1\%), stale-state reading (13.2\%), incomplete answers (12.7\%), and fact-extraction distortion (11.8\%); generation hallucination accounts for 8.5\%. \textsc{Build} failures mainly indicate that key facts were not written completely or accurately, whereas \textsc{Read} failures more often reflect obsolete-version selection or insufficient answer coverage. No entity-linking or Wiki state-location errors are observed, suggesting stable entity routing and deterministic addressing. Overall, the main bottleneck has shifted from retrieval recall to offline write fidelity, complex state updates, and answer completeness during reading.

\subsection{State Flips and Evidence Position}

Figure~\ref{fig:state_flips} groups CrossMeet-EN questions by entity-state flips; here and in subsequent analyses, VanillaRAG denotes the VanillaRAG (Hybrid) variant in Table~\ref{tab:main_results}. As flips rise from 0--1 to four, EvoWiki declines only from 71.03\% to 66.90\%, a drop of 4.13 percentage points. Direct LLM falls from 70.62\% to 61.20\%, while VanillaRAG, GraphRAG, and LogicRAG fall from 64.57\%, 68.93\%, and 65.18\% to 42.80\%, 46.80\%, and 45.32\%. On the high-conflict four-flip subset, EvoWiki leads these methods by 5.70, 24.10, 20.10, and 21.58 points, showing that explicit state overwriting limits degradation as version conflicts accumulate.

Figure~\ref{fig:position_effect} divides the 470 answerable LongMemEval instances into five evidence-position bins. EvoWiki remains between 63.58\% and 70.10\%, with a spread of 6.52 points. In the middle bin, Direct LLM, VanillaRAG, GraphRAG, and LogicRAG score 11.36\%, 30.93\%, 32.12\%, and 43.24\%, whereas EvoWiki reaches 63.58\%, leading the runner-up by 20.34 points. Because \textsc{Read} operates on a Wiki reorganized by entity and state, answers no longer depend on evidence position in the original meeting sequence, effectively mitigating the Lost-in-the-Middle effect \citep{liu2024lost}.

\subsection{Qualitative Case Study}

Table~\ref{tab:appendix_case_study_risk} presents cross-meeting risk evolution and final-state recovery. Nexus Analytics' primary risk changes from a SOC2 certification delay to insufficient customer-onboarding capacity and is ultimately overwritten by the risk that the third customer's contract negotiation may slip beyond Q4 and affect the ARR milestone. EvoWiki returns the final valid risk. Direct LLM and GraphRAG remain at the intermediate Meeting 4 state; VanillaRAG returns the SOC2 risk, already superseded twice; and LogicRAG abstains incorrectly. This case shows that semantically relevant evidence need not be the currently valid state and highlights the role of state overwriting and provenance chains.

\subsection{Human Evaluation}

Beyond automatic metrics, three researchers with NLP backgrounds independently conduct blinded evaluations on the same 200 CrossMeet-EN questions: for each question, EvoWiki's answer is paired separately with anonymized answers from Direct LLM, VanillaRAG, GraphRAG, and LogicRAG. Unaware of method identity, they judge EvoWiki as a win, tie, or loss on faithfulness, completeness, and conciseness. Final labels follow majority vote, with three-way disagreements resolved through discussion. Figure~\ref{fig:human_eval_wtl} reports results from EvoWiki's perspective. Against the four baselines, EvoWiki's faithfulness win rates are 31.0\%--51.5\%, above loss rates of 21.0\%--26.5\%. Against the three RAG methods, completeness win rates are 51.5\%--62.5\%, above loss rates of 16.0\%--23.0\%, while EvoWiki ties Direct LLM. For conciseness, its win rates against Direct LLM, VanillaRAG, and GraphRAG are 48.5\%, 43.0\%, and 42.0\%, above loss rates of 27.5\%, 34.5\%, and 32.0\%; against LogicRAG, the 28.5\% win rate is slightly below the 32.0\% loss rate. Human evaluation shows that the advantage arises from factual faithfulness and information completeness rather than generation verbosity.

\subsection{Abstention}

\begin{table}[t]
    \centering
    \small
    \resizebox{\columnwidth}{!}{
        \begin{tabular}{@{} l c c c @{}}
            \toprule
            \textbf{Method} & \textbf{CM-ZH} & \textbf{CM-EN} & \textbf{LongMemEval} \\
            \midrule
            Direct LLM           & \textbf{1.0\%} & 0.3\%  & 70.6\% \\
            VanillaRAG (BM25)    & 3.6\%  & 8.6\%  & 54.9\% \\
            VanillaRAG (Dense)   & 2.5\%  & 4.1\%  & 55.7\% \\
            VanillaRAG (Hybrid)  & 2.3\%  & 3.3\%  & 54.5\% \\
            RAPTOR               & 3.7\%  & 10.7\% & 52.6\% \\
            GraphRAG             & 1.3\%  & 1.8\%  & 51.7\% \\
            HippoRAG2            & 3.1\%  & 11.7\% & 54.5\% \\
            LightRAG             & 2.1\%  & 4.8\%  & 46.8\% \\
            LogicRAG             & 10.1\% & 17.8\% & 34.5\% \\
            EvoWiki              & 1.1\%  & \textbf{0.2\%}  & \textbf{22.8\%} \\
            \bottomrule
        \end{tabular}
    }
    \caption{Abstention rates on CrossMeet and LongMemEval.}
    \label{tab:single_column_results}
\end{table}

Table~\ref{tab:single_column_results} compares abstention rates. Every CrossMeet question is answerable, so abstention is erroneous. EvoWiki's rates are 0.2\% on CrossMeet-EN and 1.1\% on CrossMeet-ZH, indicating that Wiki-only \textsc{Read} usually provides sufficient valid states. Of LongMemEval's 500 instances, 30 are unanswerable. To make abstention reflect failure to recover a valid state, we compute it only on the remaining 470 answerable instances, where any abstention is erroneous. EvoWiki's rate is 22.8\%, below LogicRAG's 34.5\%, VanillaRAG's 54.5\%, and Direct LLM's 70.6\%.

\section{Conclusion}

We presented EvoWiki for dynamic cross-meeting QA, combining incremental \textsc{Build} and Wiki-only \textsc{Read} with entity version chains, write-time coreference resolution, state overwriting, and meeting-level provenance to maintain a current view and traceable history. We also introduced bilingual CrossMeet for factual consistency, temporal reasoning, and multi-hop QA. Across six datasets and two readers, EvoWiki achieves macro-average Judge Accuracy of 60.09 and 63.02, surpassing the strongest baselines by 9.72 and 10.00 points; state-flip, evidence-position, and human analyses confirm robust current-state reading and traceability. Wiki-only \textsc{Read} remains constrained by \textsc{Build} extraction completeness, especially under ASR noise and informal speech. Future work will explore uncertainty-aware writing, selective source verification, and broader languages, domains, and time horizons.

\bibliography{aaai2027}

\end{document}